\documentclass[10pt,twocolumn,letterpaper]{article}

\usepackage{cvpr}              

\usepackage[dvipsnames]{xcolor}
\usepackage{times}
\usepackage{epsfig}
\usepackage{graphicx}
\usepackage{amsmath}
\usepackage{amssymb}
\usepackage{algorithm}
\usepackage{algorithmicx}
\usepackage{algpseudocode}
\usepackage{multirow}
\usepackage{booktabs}
\usepackage{verbatim}
\usepackage{bm}
\usepackage{enumitem}
\usepackage{pifont}
\usepackage[accsupp]{axessibility}
\usepackage{multirow}
\usepackage{makecell}
\usepackage{adjustbox}
\usepackage{bbding}
\usepackage{tikz}
\usepackage{pgfplots}
\usepackage{ulem}
\usepackage{mathtools}
\usepackage{pgfplotstable}
\usepackage{cite}

\usepackage[pagebackref,breaklinks,colorlinks,citecolor=cvprblue]{hyperref}
\definecolor{cvprblue}{rgb}{0.21,0.49,0.74}

\usepackage[capitalize]{cleveref}
\usepgfplotslibrary{groupplots}
\usetikzlibrary{patterns}

\crefname{section}{Sec.}{Secs.}
\Crefname{section}{Section}{Sections}
\Crefname{table}{Table}{Tables}
\crefname{table}{Tab.}{Tabs.}

\newcommand{\cmark}{\ding{51}}%
\newcommand{\xmark}{\ding{55}}%

\definecolor{Gray}{gray}{0.9}
\definecolor{color1}{rgb}{0, 0.4470, 0.7410}
\definecolor{color2}{rgb}{0.8500, 0.3250, 0.0980}
\definecolor{color3}{rgb}{0.9290, 0.6940, 0.1250}
\definecolor{color4}{rgb}{0.4940, 0.1840, 0.5560}

\newcommand\blinewidth{2.0pt}

\definecolor{ms_note}{RGB}{0, 181, 190}

\def\paperID{13122 \vspace{-1em}} 
\def\confName{CVPR}
\def\confYear{2024}

\title{An Empirical Study of the Generalization Ability\\ of Lidar 3D Object Detectors to Unseen Domains  \vspace{-1em}}

\author{George Eskandar$^{1}$, Chongzhe Zhang$^{1}$, Abhishek Kaushik$^{1}$, Karim Guirguis$^{2}$, Mohamed Sayed$^{3}$, Bin Yang$^{1}$\\
{ University of Stuttgart, Germany}$^1$ \hspace{0.2em} { Karlsruhe Institute of Technology, Germany}$^2$ \hspace{0.2em} { Niantic, UK}$^3$ \hspace{0.2em}
}

\begin{document}
\maketitle
\begin{abstract}
\vspace{-1em}
3D Object Detectors (3D-OD) are crucial for understanding the environment in many robotic tasks, especially autonomous driving. Including 3D information via Lidar sensors improves accuracy greatly. However, such detectors perform poorly on domains they were not trained on, i.e. different locations, sensors, weather, etc., limiting their reliability in safety-critical applications. There exist methods to adapt 3D-ODs to these domains; however, these methods treat 3D-ODs as a black box, neglecting underlying architectural decisions and source-domain training strategies. Instead, we dive deep into the details of 3D-ODs, focusing our efforts on fundamental factors that influence robustness prior to domain adaptation.

We systematically investigate four design choices (and the interplay between them) often overlooked in 3D-OD robustness and domain adaptation: architecture, voxel encoding, data augmentations, and anchor strategies. We assess their impact on the robustness of nine state-of-the-art 3D-ODs across six benchmarks encompassing three types of domain gaps - sensor type, weather, and location.

Our main findings are: (1) transformer backbones with local point features are more robust than 3D CNNs, (2) test-time anchor size adjustment is crucial for adaptation across geographical locations, significantly boosting scores without retraining, (3) source-domain augmentations allow the model to generalize to low-resolution sensors, and (4) surprisingly, robustness to bad weather is improved when training directly on more clean weather data than on training with bad weather data. We outline our main conclusions and findings to provide practical guidance on developing more robust 3D-ODs.
   
\end{abstract}

\vspace{-1em}
\section{Introduction}
\label{sec:intro}

\begin{figure}[t!]
    \centering
    \includegraphics[width=0.98\linewidth]{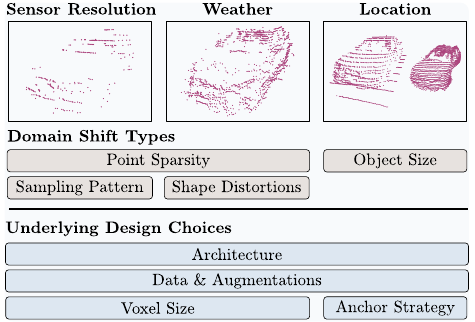}
    \caption{An overview of the three main domain gaps under study. We show the main shift types for each domain gap along with the underlying design choices that influence model robustness.}
    \label{fig:teaser}
    \vspace{-2em}
\end{figure}


The key objective of 3D object detection (3D-OD) is to accurately localize and identify objects of different classes in the 3D environment using sensor data such as point clouds~\cite{pvrcnn, second, pointrcnn} or images~\cite{fcos3d, bevdet, bevformer}. Lidar sensors have played a pivotal role in this endeavor, as they can provide dense and highly precise point clouds using a long scanning range. While 3D object detection models reliant on Lidar point clouds have achieved remarkable results on demanding benchmarks such as KITTI~\cite{kitti}, Waymo~\cite{waymo}, and NuScenes~\cite{nuscenes}, they face a significant performance drop when deployed in unfamiliar settings. This decline in performance might arise from migrating to a different sensor, owing to changes in point sparsity levels and scanning patterns, or from changes in the perceived 3D shapes of objects due to adverse weather conditions like rain, fog, or snow. Other changes can stem from deployment in a new geographical location that exhibits different statistical properties like smaller or bigger object sizes. All these scenarios pose a considerable threat to the safety and reliability of autonomous driving. 

Domain Adaptation (DA) promises to address these issues by adapting models trained on a source domain to a target domain, achieving strong results on some domain gaps~\cite{sn, spg, st3d, st3d++, 3dcoco, jst, mlc, umt, ssda3d, beamdistillation, sailor, sourcefree, dataadaption, atsim2realOD}. However, these methods often treat detection models as black-boxes relying heavily on the availability of target domain data. This leaves a gap in understanding: \textit{how do architectural choices and training strategies affect generalization to target domains?} Furthermore, not all DA methods develop their techniques using the same architectures, making it even harder to evaluate these models. Unlike images, point clouds are unordered 3D structures, allowing for a broader range of Lidar-based detector design options - operating on points, voxels, or a combination of both.

Understanding the robustness of 3D Lidar detectors thus holds significant importance, given that data distribution shifts are inevitable and more likely to occur with the growing deployment of autonomous vehicles. Recently, there have been studies in this field~\cite{understandingbev, bench_corruptions, bench_fusion, weather_robustness}, focusing primarily on corruptions benchmarking and identifying sensor representations (RGB, Lidar, or fusion) exhibiting greater resilience. Notably, these works highlight that fusion models suffer greater when the Lidar input, as opposed to RGB, is corrupted. Following this, our particular focus lies on Lidar-only 3D-ODs and the impact of design choices on detection robustness, ultimately providing recommendations on standard practice and solutions. Furthermore, unlike previous works~\cite{understandingbev, bench_corruptions}, we study domain gaps that are not necessarily caused by environmental or sensor corruptions but are still harmful to the detection performance, such as different geo-locations and sensor resolution. We methodically isolate and study each domain gap when benchmarking, giving careful attention to the differences in sensor differences, location, and weather.

\noindent\textbf{Contribution.} In this work, we methodically study the effect of several design choices in the architecture and training strategy on the generalization of Lidar-based detectors to unseen domains. Our aim is to preserve as much accuracy as we can before relying on domain adaptation. We present a taxonomy of various domain shift types in each domain gap (\cref{fig:teaser}) and four underlying design choices (architecture, augmentation, voxel size, anchor size) that can potentially address these domain shifts. Then, in Sec.~\ref{sec:benchmarking}, we evaluate nine state-of-the-art 3D detectors on six DA benchmarks featuring three domain gaps (weather, location, and resolution), analyzing the impact of many common architectural designs. A large number of controlled experiments and apple-to-apple comparisons are subsequently conducted to disentangle the best practices for each domain gap in sections \ref{sec:location}, \ref{sec:sensor} and \ref{sec:weather}. Similar to previous devil-in-the-detail investigative works for CNNs on images~\cite{devil, devil2, delving, blur}, we show that often simple and overlooked details can significantly influence out-of-domain (OOD) performance. The main focus of this work is to provide an empirical study to understand the robustness of different design choices in Lidar 3D-OD. We also provide solutions to some of the problems based on our findings and point out potential research areas. We present a summary of the novel findings and practical recommendations in Sec.~\ref{sec:conclusion}.

\section{Related Works}
\noindent\textbf{Robustness Benchmarks for 3D-OD} is a relatively new research area that seeks to study the robustness of 3D detectors under different unseen corruptions. Previous studies~\cite{bench_fusion, bench_corruptions, understandingbev} primarily concentrated on categorizing these corruptions and establishing benchmarks to evaluate the OOD performance of different sensor representations (camera, Lidar, fusion of both). \cite{bench_fusion} introduced a benchmark for camera-lidar fusion models, revealing a heavy reliance on Lidar as these models fail worse when only Lidar data is corrupted. \cite{bench_corruptions} established more benchmarks showing that fusion models are the most robust, while the camera-only models are the most vulnerable. Both studies underscore the importance of Lidar. Finally, \cite{understandingbev} studied the robustness of bird-eye view representations in camera-only and fusion models when subjected to environmental corruption and adversarial attacks. Motivated by prior research, our study focuses exclusively on Lidar-only models and delves into the intricate relationship between design choices in the 3D-OD pipeline and robustness in unseen domains. While this has been extensively studied in images~\cite{benchmarking, ibnnet, danorm, daformer}, it is still lacking in Lidar.

\begin{table*}[th!]
\centering
\renewcommand\arraystretch{0.95}
\setlength{\tabcolsep}{10pt}
\adjustbox{width=0.9\linewidth}{
\begin{tabular}{c|c|cc|c|c|c}
\toprule
\multirow{2}{*}{Representation} & \multirow{2}{*}{Model} & \multicolumn{2}{c|}{3D Feature Encoder} & BEV & \multirow{2}{*}{DenseHead} & \multirow{2}{*}{RCNN} \\
& & Voxels & Points & Backbone & \\
\Xhline{1pt}

Points & PointRCNN~\cite{pointrcnn} & \xmark & PointNet & \xmark & PointHead & PointRCNN\\
\midrule
\multirow{5}{*}{Voxels} & PointPillars~\cite{pointpillars} & \xmark & \xmark & \cmark & AnchorHead  & \xmark \\
& Second~\cite{second} & 3DCNN$\times8$ & \xmark & \cmark & AnchorHead & \xmark \\
& Voxel-RCNN~\cite{voxelrcnn} & 3DCNN$\times8$ & \xmark & \cmark & AnchorHead & Voxel \\
& VOTR-VoxelRCNN & VOTR & \xmark & \cmark& AnchorHead & Voxel \\
\midrule
\multirow{3}{*}{Hybrid (PV)} & CenterPoint~\cite{centerpoint} & 3DCNN$\times8$ & \xmark & \cmark & Centerhead & Point \\
& PV-RCNN-Centerhead~\cite{openpcdet} & 3DCNN$\times8$ & VSA & \cmark & CenterHead & PV \\
& PV-RCNN~\cite{pvrcnn} & 3DCNN$\times8$ & VSA & \cmark & AnchorHead & PV \\
&VOTR-TSD~\cite{votr} & VOTR & VSA & \cmark & AnchorHead & PV\\
\bottomrule
\end{tabular}}
\vspace{-0.5em}
\caption{Taxonomy of state-of-the-art Lidar 3D detectors based on their common and specific architectural components. }
\label{tab:model_zoo}
\vspace{-1em}
\end{table*}


\noindent\textbf{Domain Adaptation for Lidar 3D-OD} has started to gain more attention in the past few years but is still not as mature as DA on images~\cite{dafaster, crossmeanteacher, htcn, i3net, epm}. SN~\cite{sn} has identified object size as the major shift type across different locations and introduced an approach to resize the source domain labels based on the mean object size (MOS) of the target domain. Many approaches have then followed, leveraging self-training methods~\cite{st3d, st3d++, jst, umt, mlc, ssda3d, beamdistillation, sourcefree}, contrastive instance-level feature alignment~\cite{3dcoco}, adversarial learning~\cite{atsim2realOD}, data augmentations~\cite{dataadaption, st3d, st3d++, beamdistillation, ssda3d} or anchor scaling~\cite{jst, sailor}. Lidar beam distillation~\cite{beamdistillation} specifically addresses the cross-resolution domain gap by learning progressively downsampled versions of the source domain with a teacher-student model while employing ST3D~\cite{st3d} on the target domain data. 3D-VField \cite{3DVField} introduces a data augmentation for synthetic-to-real domain generalization, where foreground objects are deformed using vector fields. To improve the generalization to adverse weather conditions, SPG \cite{spg} restores missing points through a self-supervised method. By simulating the characteristics of lasers in foggy settings, \cite{fog} transforms pointclouds collected in sunny scenes into pointclouds in fog and use them as data augmentation.

\section{Benchmarking}
\label{sec:benchmarking}

It has been shown in several works on image perception~\cite{daformer, benchmarking, intriguing, robustvit, texturebiascnn} that the network's architecture has a significant impact on its generalization ability. Here, we seek to answer this question in the context of Lidar 3D-OD. 

\subsection{Architectures}
We take a closer look at nine state-of-the-art 3D detectors (using their official code based on OpenPCDet~\cite{openpcdet}) and present a taxonomy of their architectural components in \cref{tab:model_zoo}. We break down the architecture of a 3D detector into four main parts: a 3D feature encoder, a 2D BEV backbone, a dense head, and an RCNN head (in the case of a two-stage model). We find that these nine frameworks can be further categorized into three main approaches based on the pointcloud representation: point-based, voxel-based, and hybrid-based (points and voxels (PV)). 

Point-based approaches like PointRCNN~\cite{pointrcnn} employ a PointNet~\cite{pointnet}-like feature extractor and an anchor-based dense head ( PointHead) processing point features. On the other hand, voxel-based approaches discretize the pointcloud into pillars and process the resulting BEV features with a 2D backbone (Pillars) or discretize into voxels and employ a 3DCNN with $8\times$ downsampling (SECOND). Two-stage voxel detectors (VoxelRCNN) employ an additional voxel-only RCNN head. We also experiment by replacing the 3DCNN in VoxelRCNN with the VOTR backbone in VOTR-TSD~\cite{votr} and name this model VOTR-VoxelRCNN. Lastly, in hybrid approaches like PVRCNN, PVRCNN-Centerhead and VOTR-TSD, point features are extracted alongside voxel features via the voxel set abstraction (VSA) module~\cite{pvrcnn}, which are then used in the RCNN head. Note that while CenterPoint uses voxels only in the feature extractor, its CenterHead also uses point features so we classify it as a hybrid approach. \cref{tab:model_zoo} can be viewed as an ablation study on a single abstract 3D object detector, allowing us to perform apple-to-apple comparisons and pinpointing the effect of single architectural components.

\subsection{Experimental Setup}

We choose six common benchmarks formed by four datasets to evaluate the robustness of the studied models. An overview of the utilized datasets and the existing domain gaps between them are shown in ~\cref{tab:datasets}. To study the resolution domain gap in more details, we follow \cite{beamdistillation} and downsample the KITTI dataset two times (KITTI-32) and four times (KITTI-16). We report the 3D Average Precision (3D AP) with 40-points Recall on the classes car, pedestrian and cyclist using the official evaluation metrics for each benchmark: KITTI metrics for Waymo-to-Kitti (W$\rightarrow$K), Waymo-to-NuScenes (W$\rightarrow$N), NuScenes-to-KITTI (N$\rightarrow$K), KITTI64-to-KITTI32 (K64$\rightarrow$K32) and KITTI64-to-KITTI16 (K64$\rightarrow$K16), and Waymo metrics for Waymo-to-Kirk (W$\rightarrow$Kr). Note that we train on the $20\%$ training split of the Waymo dataset to achieve a high number of experiments. In all experiments, we choose the best-performing checkpoint on the validation set of the source domain and evaluate it on the target domain validation split, following DA works~\cite{st3d}.

\begin{table}[!t]
    \centering
    \setlength{\tabcolsep}{1pt}
    \adjustbox{width=\linewidth}{\begin{tabular}{c|c|c|c|c|c}
    \toprule
    Dataset & VFOV & Lines & \#Samples & Location & Weather \\   
    \hline
    KITTI & $[-23.6^{\circ}, 3.2^{\circ}]$ & 64 & 14k/2k/734 & Germany & Clear \\
    Waymo & $[-17.6^{\circ}, 2.4^{\circ}]$ & 64 & 4.7M/2.2M/53k & USA & Clear  \\
    Kirkland & $[-17.6^{\circ}, 2.4^{\circ}]$ & 64 & 312k/21k/0 & Kirkland & Rainy  \\
    nuScenes & $[-30.0^{\circ}, 10.0^{\circ}]$ & 32 & 196k/100k/10k & Boston \& Singapore & Clear  \\
    \bottomrule
    \end{tabular}}
    \vspace{-0.5em}
    \caption{A summary of the datasets used in this study. \# Training samples is reported for Car/Pedestrian/Cyclist.}
    \label{tab:datasets}
    \vspace{-1.5em}
\end{table}
\begin{table*}[!ht]
    \centering
    \renewcommand\arraystretch{1}
    \adjustbox{width=\linewidth}{
    \begin{tabular}{c|c|c|ccc|ccc|ccc|ccc|cc|ccc|ccc}
    \hline
    \toprule
        \multirow{2}{*}{Features} & \multirow{2}{*}{Architecture} & \multirow{2}{*}{Method} & \multicolumn{3}{c|}{K64$\rightarrow$K32 (R)} & \multicolumn{3}{c|}{K64$\rightarrow$K16 (R)} & \multicolumn{3}{c|}{W$\rightarrow$K (G)}  & \multicolumn{3}{c|}{W$\rightarrow$N (G+R)} & \multicolumn{2}{c|}{W$\rightarrow$Kr (W)} & \multicolumn{3}{c|}{N$\rightarrow$K (G+R)} & \multicolumn{3}{c}{Mean} \\
        ~ & ~ & ~ & Car & Ped & Cyc & Car & Ped & Cyc  & Car & Ped & Cyc  & Car & Ped & Cyc  & Car & Ped & Car & Ped & Cyc  & Car & Ped & Cyc  \\
        
        \midrule
        
        Point & MLP & PointRCNN & 74.7 & 51.86 & \textbf{61.52} & 51.74 & 21.03 & 24.71 & 5.04 & 28.22 & 0.0 & 12 & 3.92 & 0.0 & 19.24 & 5.86 & 13.81 & 25.31 & 0.0 & 29.42 & 22.7 & 18.25  \\ 
        
        \midrule
        
        Voxel & Conv & PointPillars & 70.48 & 35.81 & 25.93 & 53.16 & 20.5 & 9.28 & 12.75 & 48.34 & 34.9 & 20.9 & 5.45 & 0.04 & 46.77 & 13.79 & 0.0 & 0.0 & 0.0 & 34.01 & 20.65 & 10.93  \\ 
        
        ~ & Conv & Second & 73.5 & 41.11 & 39.1 & 50.71 & 16.63 & 17.58 & 9.91 & 41.39 & 22.74 & 17.84 & 4.44 & 0.22 & 46.04 & 14.97 & 6.36 & 13.43 & 0.0 & 34.06 & 22 & 15.17 \\ 
        
        ~ & Conv & VoxelRCNN & \textbf{76.96} & \textbf{56.71} & 50.18 & \textbf{56.11} & \textbf{27.24} & 24.64 & \textbf{20.09} & \textbf{55.33} & 34.81 & 19.71 & 0.05 & 0.0 & 52.18 & 20.7 & 7.48 & 20.96 & 0.0 & 38.75 & \textbf{30.16} & 19.47   \\ 
        
        ~ & ViT & VOTR-VoxelRCNN &76.05 & 51.62 & 46.04 & 55.7 & 24.55 & 20.79 & 21.34 & 29.96 & 1.63 & 15.34 & 0.02 & 0.0 & 49.21 & 15.81 & 19.26 & \textbf{26.82} & 2.04 & 39.48 & 24.8 & 22.48\\ 
        
        \midrule 
        
        Hybrid & Conv & CenterPoint & 71.18 & 42.71 & 44.19 & 52.8 & 15.11 & 14.27 & 13.78 & 53.39 & \textbf{42.43} & 19.05 & 5.63 & 0.5 & 50.23 & \textbf{30.72} & 8.96 & 19.94 & 0.04 & 36 & 27.92 & 18.57 \\ 
        
        ~ & Conv & PVRCNN Centerhead & 71.39 & 44.64 & 34.2 & 44.24 & 15.39 & 12.02 & 15.06 & 52.47 & 40.13 & 20.59 & 6.84 & 0.42 & 52.79 & 29.02 & \textbf{28.52} & 21.67 & 0.03 & 38.76 & 28.34 & 23.46 \\ 
        
        ~ & Conv & PVRCNN & 77.62 & 53.96 & 50.3 & 54.07 & 27.13 & \textbf{26.36} & 16.61 & 50.22 & 34.09 & 20.36 & 5.79 & 0.53 & \textbf{54.34} & 25.79 & 10.46 & 17.46 & 0.0 & 38.91 & 30.06 & 19.38 \\ 
        
        ~ & ViT & VOTR-TSD & 76.29 & 49.78 & 49.76 & 55.61 & 21.71 & 24.38 & 15.75 & 46.19 & 39.28 & \textbf{21.32} & \textbf{7.0} & \textbf{3.48} & 52.52 & 26.56 & 26.26 & 26.56 & \textbf{4.7} & \textbf{41.29} & 29.63 & \textbf{25.69 } \\ 
        
        \bottomrule
    \end{tabular}}
    \vspace{-0.5em}
    \caption{Evaluation of different architectures on six DA benchmarks. Beside each benchmark, we denote whether it is mainly caused by a resolution (R), weather (W), or geographical location (G) discrepancy. Voxel Transformers and hybrid representations are found to be more robust across, on average, across the considered domains. Note that the Kirk dataset does not have the class cyclist.}
    \label{tab:benchmark}
    \vspace{-1em}
\end{table*}

\subsection{Results}
We report the results in \cref{tab:benchmark} and the key findings.


Finding 1: \textit{VOTRs outperform 3D CNNs in the mean AP across all domain gaps when coupled with point features (VOTR-TSD).} This finding only partially aligns with observations in image perception literature~\cite{texturebiascnn, robustvit, daformer}, where it has been shown that transformers are more robust than 3D CNNs. We observe that 3D CNNs can be more robust than VOTRs in voxel-only models (VoxelRCNN versus VOTR-VoxelRCNN). The addition of point features enhances the performance of VOTR significantly, as it adds a much-needed spatial local context to the transformer's large receptive field of view. This is more important in Lidar than images, because the relative size of some classes, like Pedestrians, to the input size is much smaller in pointclouds than in images.


Finding 2: \textit{Anchorless detectors are robust in the weather domain gap.} This is more evident on the Pedestrian class, where CenterPoint and PVRCNN-Centerhead outperform other models by 4-5 AP.

Finding 3: \textit{Adding point features in the backbone increases robustness, especially for transformers.} VOTR-TSD outperforms VOTR-VoxelRCNN (Tab.~\ref{tab:benchmark}), PVRCNN-Centerhead outperforms Centerpoint but PVRCNN and VoxelRCNN have similar performance. The key distinction between these pairs is point features in the backbone. 


\section{Location Domain Gap}
\label{sec:location}
It has been shown that discrepancies in object sizes across geographical locations are the main cause of bad generalization~\cite{sn, st3d, st3d++}. Explanations and remedies to this problem are controversial: while some works address this problem by varying the label size on the source domain~\cite{st3d, sn}, others~\cite{sailor, jst} change the anchor size without a labeled target set with mixed results. As both strategies were explored in different settings, it is not clear which is the most beneficial. In this section, we perform experiments on the W$\rightarrow$K, W$\rightarrow$N and N$\rightarrow$K benchmarks to highlight the most influential design choices. We train exclusively on the class car as it exhibits the most variations across different locations, following~\cite{sn, st3d} (see Appendix for Multiclass experiments).

\subsection{Anchor Size} The anchor size in 3D-OD is a carefully tuned hyperparameter and is usually close to the mean object size (MOS) of the training dataset. In \cref{fig:anchor_size}, we seek to answer two questions: Is there a connection between anchor size and performance in OOD scenarios? And, when considering training and test phases, which holds greater significance: the anchor size during training or testing? We perform our experiments using the SECOND model~\cite{second} due to its good performance in DA benchmarks~\cite{st3d} and its low computation requirements. We train three models, each with a different anchor. The first uses the default training anchor on Waymo dataset, while the second and third use a smaller and a larger one, respectively. We notice the following: 1) \textit{There is no correlation between the anchor size at training time and the OOD performance}. All three models have low AP when they use the same anchor size they were trained on. 2) However, \textit{changing anchor size at test time leads to strong variation in the performance}. 3) \textit{The best-performing anchors are found to be always smaller than the training anchor}. This is also true when the target MOS is bigger than the source MOS (see Appendix). 4) There exists indeed a \textit{test-time anchor size that yields a very high performance} on the target domain without retraining. This is heuristically determined with a simple greedy algorithm: we change one dimension at a time and select the value that achieves the best AP on the validation set. Then, we fix this value and change the subsequent dimensions following the same procedure (see Appendix). While this is the best anchor we find using this optimization procedure, there might be a more optimal solution. Finally, note that the described technique of changing anchor size at test-time is \textit{not meant} to be a proposed domain generalization approach (since we directly tune the anchor on the target dataset). Instead, we reveal that this simple trick can significantly enhance the performance in a semi-supervised, weak, or federated domain adaptation setting. 
\begin{figure}[t!]
    \centering
    \includegraphics[width=0.95\linewidth]{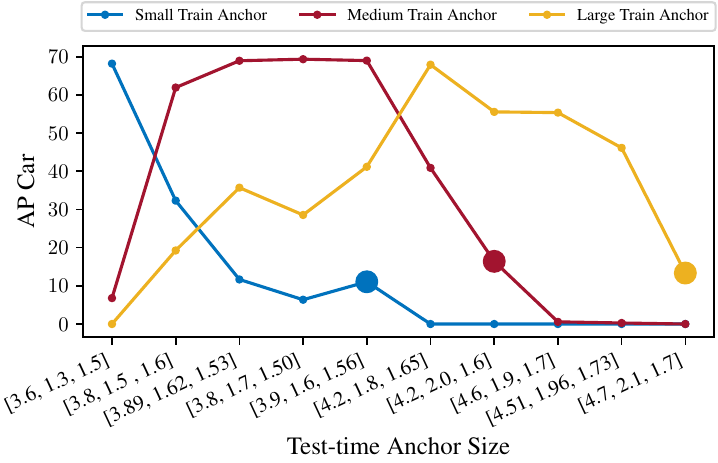}
    \vspace{-0.5em}
    \caption{Evaluation of different anchor sizes (in the order of increasing volumes) at test-time on W$\rightarrow$K car benchmark. Three training experiments using SECOND are performed, each with a different anchor size. Using the same size at test-time results in a poor performance ($\circ$), but a very high peak can be obtained by going for lower anchor sizes.}
    \label{fig:anchor_size}
    \vspace{-2em}
\end{figure}
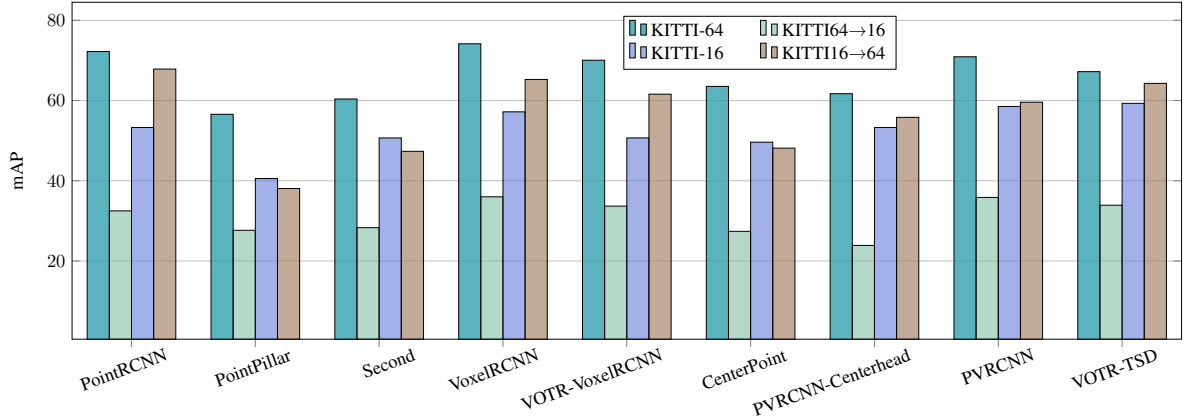
\begin{figure*}[t!]     
\centering
 \begin{adjustbox}{width=0.9\textwidth}
   \begin{tikzpicture}

\begin{axis}[
    ybar,
    width=1.3\textwidth,
    height=8cm,
    enlargelimits=0.06,
    legend style={
        at={(0.620,0.8)},
        anchor=south,
        legend columns=2,
        column sep=0.01cm,
        text width=2.0cm,
        /tikz/every even column/.append style={column sep=0.01cm}
    },
    ylabel={mAP},
    ylabel style={yshift=-5pt},
    symbolic x coords={PointRCNN,PointPillar,Second,VoxelRCNN,VOTR-VoxelRCNN,CenterPoint,PVRCNN-Centerhead,PVRCNN,VOTR-TSD},
    xtick=data,
    xtick pos = bottom,
    x tick label style={rotate=20,anchor=east,xshift=20,yshift=-10},
    width=1.3\textwidth,
    bar width=12pt,
    ymajorgrids=true,
    ymin=5,
    ymax=80, 
]
\addplot+[bar shift=-18pt, draw=black, fill={rgb,255:red,50; green,160; blue,175}, fill opacity=0.8] coordinates {(PointRCNN,72.22) (PointPillar,56.57) (Second,60.39) (VoxelRCNN,74.15) (VOTR-VoxelRCNN,70.05) (CenterPoint,63.53) (PVRCNN-Centerhead,61.7) (PVRCNN,70.91) (VOTR-TSD,67.2)};

\addplot+[bar shift=-6pt, draw=black, fill={rgb,255:red,107; green,180; blue,147}, fill opacity=0.5] coordinates {(PointRCNN,32.49) (PointPillar,27.65) (Second,28.3) (VoxelRCNN,36) (VOTR-VoxelRCNN,33.68) (CenterPoint,27.39) (PVRCNN-Centerhead,23.88) (PVRCNN,35.85) (VOTR-TSD,33.9)};

\addplot+[bar shift=6pt, draw=black, fill={rgb,255:red,100; green,125; blue,219}, fill opacity=0.6] coordinates {(PointRCNN,53.28) (PointPillar,40.56) (Second,50.67) (VoxelRCNN,57.18) (VOTR-VoxelRCNN,50.68) (CenterPoint,49.62) (PVRCNN-Centerhead,53.29) (PVRCNN,58.52) (VOTR-TSD,59.3)};

\addplot+[bar shift=18pt, draw=black, fill={rgb,255:red,156; green,116; blue,88}, fill opacity=0.6] coordinates {(PointRCNN,67.85) (PointPillar,38.06) (Second,47.34) (VoxelRCNN,65.26) (VOTR-VoxelRCNN,61.59) (CenterPoint,48.13) (PVRCNN-Centerhead,55.81) (PVRCNN,59.59) (VOTR-TSD,64.28)};

\legend{KITTI-64, KITTI64→16, KITTI-16, KITTI16→64}
\end{axis}

\end{tikzpicture}


\end{adjustbox}
    \vspace{-1em}
    \caption{Evaluation of all models on K64, K16, K64$\rightarrow$K16 and K16$\rightarrow$K64 benchmarks. The mAP for all classes is reported. The performance drops significantly from high-to-low resolution while it increases from low-to-high at test-time.}
    \label{fig:sensor_gap_understanding}
    \vspace{-1em}
\end{figure*}

\subsection{Anchor vs. Label Strategy} In Tab.~\ref{tab:anchorvslabel}, we contrast the two strategies against each other at training time. Specifically, we train SECOND using the random object scaling (ROS) proposed in ST3D~\cite{st3d}, where the foreground objects and their labels are randomly resized. Moreover, we train SECOND with multiple anchors (three) and include an anchorless detector in the analysis (CenterPoint). While training with anchors of different sizes is very common in 2D-OD, it is surprisingly not used in 3D-OD~\cite{openpcdet}, as models are usually trained with two anchors per class of the same size with different orientations. We find that: 1) \textit{Anchorless detectors exhibit only a marginal improvement in generalization compared to anchor-based detectors} when dealing with objects of various sizes. Their performance is notably suboptimal. 2) \textit{ROS enhances robustness on KITTI but decreases performance on NuScenes and the source domain.} 3) \textit{The utilization of multiple anchor sizes adds positive gains on Waymo and W$\rightarrow$N.} In summary, there is no universally generalizable technique at training time across all locations. From a continual learning perspective, we argue it is more beneficial to fix the original training labels and change the anchor at test time using a small set of labeled target data.   
\begin{table}[t!]
    \centering
    \adjustbox{width=0.8\linewidth}{\begin{tabular}{l|c|c|c}
        \toprule
        Model & Waymo  & W$\rightarrow$K  & W$\rightarrow$N \\ 
        \hline
        Second & 57.42 &  16.41 &  18.69 \\ 
        Second w/ ROS & 49.01 & \textbf{41.91} & 14.09 \\ 
        Second 3 Anchors & \textbf{59.86} &14.58 & \textbf{21.22} \\ 
        Second 3 Anchors w/ ROS  & 51.83 & 33.74 & 16.01 \\ 
        \hline
        CenterPoint & 58.48 & 13.86 &  20.08 \\ 
        CenterPoint w/ ROS  & 53.30 & 34.46 & 16.61 \\ 
        \bottomrule
    \end{tabular}}
    \vspace{-0.5em}
    \caption{Investigating the effect of ROS, using multiple anchors and anchorless detectors on the OOD performance of the class car. The source AP (Waymo) is also considered.}
    \label{tab:anchorvslabel}
\end{table}
\begin{table}[!t]
    \centering   
    \adjustbox{width=0.8\linewidth}{\begin{tabular}{c|c|c|c}
    \toprule
          Model & W$\rightarrow$K & W$\rightarrow$N & N$\rightarrow$K \\ \hline
          
         PointRCNN &  5.38 / 35.55  & 1.31 / 14.29 & 15.0 / 27.11 \\ 
         
         PointPillar & 10.48 / 65.14 & 21.01 / \textbf{23.84} & 0.04 / 0.01 \\

         SECOND & 16.41 / \textbf{66.81} & 18.69 / 21.31 & 4.97 / 41.64 \\
         
         VoxelRCNN & 18.28 / 59.49 & 19.47 / 21.34 & 8.74 / 29.0  \\ 
         
         VOTR-VORCNN & \textbf{18.15} / 65.14 & 19.38 / 21.18 & 18.47 / \textbf{46.96} \\
         
         PVRCNN & 9.69 / 40.86 & 20.08 / 22.31 & 14.84 / 26.67 \\ 
         
         VOTR-TSD & 14.47 / 52.64 & \textbf{21.66} / 23.74 & \textbf{24.75} / 35.0 \\
         
        \bottomrule
    \end{tabular}}
    \vspace{-0.5em}
    \caption{Benchmarking the effect of changing the anchor size on different anchor-based models: we report the 3D AP on the class car before/after tuning the anchor at test time.}
    \label{table:anchor_all_models}
    \vspace{-1em}
\end{table}


\subsection{Effects on different Architectures}
In Tab.~\ref{table:anchor_all_models}, we evaluate the effect of tuning the anchor size at test time on seven state-of-the-art anchor-based models. Results are reported on the three DA benchmarks exhibiting different object sizes: W$\rightarrow$K, W$\rightarrow$N, and N$\rightarrow$K. We find that: 1) \textit{The performance of all models increases on all 3 benchmarks}. 2) \textit{The effect is more pronounced when the discrepancy between object sizes in the source and target datasets is great.} For instance, the effect is stronger on W$\rightarrow$K than on W$\rightarrow$N. 3) \textit{This technique more positively influences voxel-only models than point-only or hybrid models}. The AP of VOTR-TSD and PVRCNN post-tuning is lower than voxel-only models. 

\section{Sensor Domain Gap}
\label{sec:sensor}

\begin{figure*}[t!] 
\centering
\begin{adjustbox}{width=0.9\textwidth}
    \centering
    \begin{tikzpicture}
\begin{groupplot}[
    group style={
        group name=my plots,
        group size=3 by 1,
        horizontal sep=1cm,
    },
    width=0.5\textwidth,
    height = 0.35\textwidth, 
    xtick=data,
    xtick pos=bottom,
    xticklabels={6.25,10,12.5},
    xticklabel style={rotate=0, anchor=east,yshift=-10pt,xshift=12pt},
    legend columns=-1,
    legend to name=globallegend,
    legend entries={PVRCNN,VOTR-TSD,SECOND},
    xlabel={Voxel Height [cm]}
    ]
\nextgroupplot[ylabel={mAP}, ytick={25, 30, 35, 40}, ymin = 23, ymax=41, title={KITTI64$\rightarrow$KITTI16}]
\addplot+[color=color1, line width=\blinewidth, mark=*, every mark/.append style={fill=color1}] coordinates {(1,33.1) (2,35.8) (3,37.8)};
\addplot+[color=color2, line width=\blinewidth, mark=square*, every mark/.append style={fill=color2}] coordinates {(1,34.37) (2,33.9) (3,38.14)};
\addplot+[color=color3, line width=\blinewidth, mark=triangle*, every mark/.append style={fill=color3}] coordinates {(1,27.91) (2,28.3) (3,30.61)};

\nextgroupplot[ytick={50, 55, 60, 65}, ymin = 48, ymax=66, title={KITTI64$\rightarrow$KITTI32}]
\addplot+[color=color1, mark=*, line width=\blinewidth, every mark/.append style={fill=color1}] coordinates {(1,59.83) (2,60.63) (3,62.83)};
\addplot+[color=color2, mark=square*, line width=\blinewidth, every mark/.append style={fill=color2}] coordinates {(1,56.71) (2,58.61) (3,61.56)};
\addplot+[color=color3, mark=triangle*, line width=\blinewidth, every mark/.append style={fill=color3}] coordinates {(1,49.91) (2,51.24) (3,52.28)};

\nextgroupplot[ytick={60, 65, 70, 75}, ymin = 58, ymax=76, title={KITTI64}]
\addplot+[color=color1, mark=*, line width=\blinewidth, every mark/.append style={fill=color1}] coordinates {(1,72.4) (2,70.8) (3,71.5)};
\addplot+[color=color2, mark=square*, line width=\blinewidth, every mark/.append style={fill=color2}] coordinates {(1,65.33) (2,67.2) (3,67.63)};
\addplot+[color=color3, mark=triangle*, line width=\blinewidth, every mark/.append style={fill=color3}] coordinates {(1,60.41) (2,60.39) (3,59.09)};
\end{groupplot}

\node at (current bounding box.north) [above, yshift=1em] {\pgfplotslegendfromname{globallegend}};
\end{tikzpicture}
\end{adjustbox}
    \vspace{-0.5em}
    \caption{ We train each model of (VOTR-TSD, PVRCNN, and SECOND) on K64 with three different voxel heights. We evaluate these nine training experiments on K64, and the lower resolutions K32, and K16. mAP for all classes is reported. Larger heights increase performance on the unseen target domains (2-7 mAP points), while keeping the source performance relatively stable (2-3 mAP points).}
    
    \label{fig:sensor_gap_voxel_height}
    \vspace{-1em}
\end{figure*}
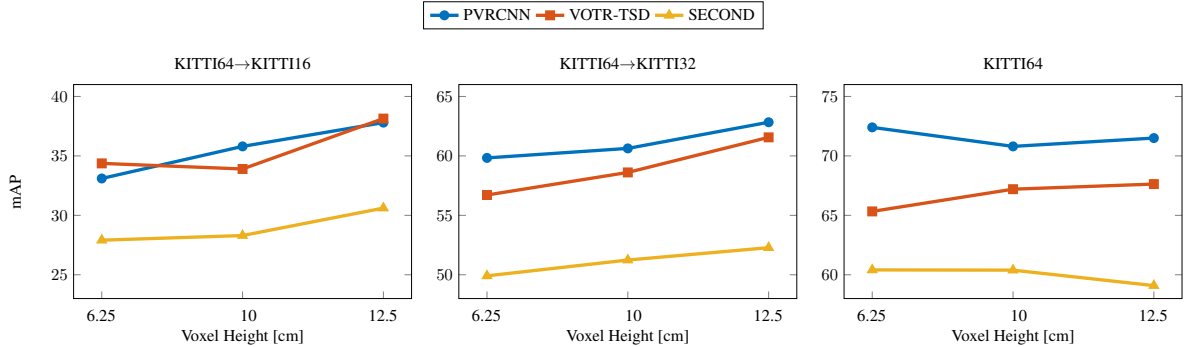

We commonly notice an underlying domain gap on the sensor level, which can sometimes lead to very different representations of the 3D environment. It is hard to isolate this domain gap since acquiring the same data with different sensors is practically troublesome. Moreover, common Lidar DA benchmarks are complicated by the fact that the gaps between source and target domains stem from multiple factors at once. For instance, the benchmarks W$\rightarrow$N and N$\rightarrow$K feature domain gaps on the sensor level and the object size. For this reason, we follow \cite{beamdistillation} and study the benchmarks which predominantly exhibit a discrepancy on the sensor level, namely K64$\rightarrow$K32, K64$\rightarrow$K16, K16$\rightarrow$K64 and W$\rightarrow$N (since the average object sizes between Waymo and NuScenes are close). We make the distinction between two cases: (1) same Vertical-FOV and different number of beams, and (2) different V-FOV. The second case is represented by W$\rightarrow$N, resulting in a different sampling pattern and is considerably harder to solve.  

\subsection{Low-to-High vs. High-to-Low}
In \cref{fig:sensor_gap_understanding}, we compare the performance of all models on the benchmarks K64$\rightarrow$K16 and add K16$\rightarrow$K64, presenting a high-to-low and low-to-high domain gaps respectively. We report the mean AP for all 3 classes (car, pedestrian and cyclist) and add oracles on K64 and K16 as well. The results consistently show that the \textit{high-to-low domain gap is much harder to tackle than the low-to-high gap}, as all models trained on K16 achieve a higher mAP on K64 (out-of-domain) than K16 (in-domain). While the performance of K16$\rightarrow$K64 is still lower than the Oracle on K64, the difference is small for most models. In the rest of this section, we focus on the High-to-Low discrepancy.

\begin{figure}[t!]
    \centering
    \includegraphics[width=0.95\linewidth]{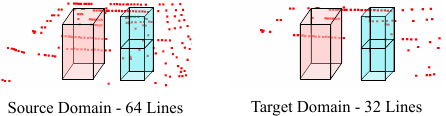}
    \caption{\textbf{Left:} Source domain with 64 lines. \textbf{Right:} In a target domain with 32 lines, a large number of voxels will be empty if the voxel height is too small (blue voxel), but this number will be reduced if the voxel height is increased (red voxel).}
    \label{fig:voxel_size}
    \vspace{-1em}
\end{figure}

\begin{figure*}
    \centering
    \includegraphics[width=\textwidth]{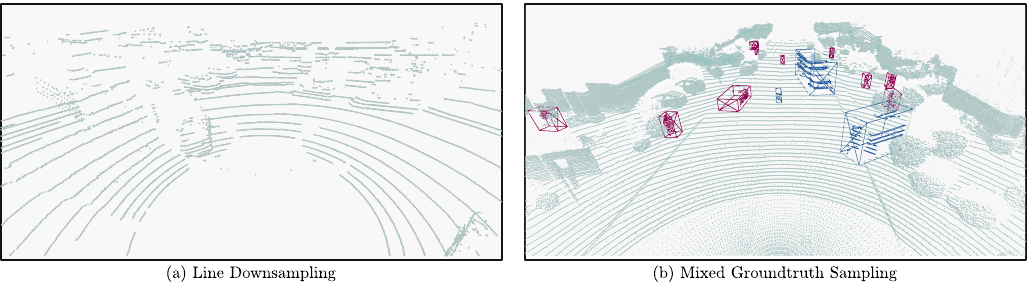}
    \caption{Visualizattion of the two introduced augmentations on the Waymo dataset. In (b), objects with different colors denote the augmented groundtruth samples from the target domain (KITTI).}
    \label{fig:vis_aug}
    \vspace{-0.5em}
\end{figure*}

\subsection{Voxel Encoding}
In \cite{second, pvrcnn}, it is observed that reducing the voxel size improves source domain performance. The impact of this change on the model's ability to handle sparser pointclouds at test time remains unclear. A common view in the field~\cite{beamdistillation} is that a smaller number of points per foreground object at test time is behind the performance drop. We reframe this explanation on the voxel level: a smaller number of beams results in more empty voxels vertically, which will not be used for detection. The smaller the voxel height, the stronger this effect becomes. On the other hand, a larger height would reduce the difference in the number of empty voxels between source and target domains. In our investigation, we train models with different voxel heights in \cref{fig:sensor_gap_voxel_height}. We validate this hypothesis on three models (SECOND, PVRCNN and VOTR-TSD), by changing the voxel height from $0.1m$ on KITTI to $0.0625m$ (small) and $0.125m$ (high). Results confirm \textit{that there is a positive correlation between voxel height and robustness on low-resolution sensors} as the 3D mAP increases in most experiments on the target domain (K32 and K16) while remaining relatively stable on the source domain (K64). The improvement is especially pronounced in the sparsest domain (K16), where models exhibit an increase of 3 to 7 mAP points. A simple conceptual illustration is drawn in \cref{fig:voxel_size}. 

\begin{table*}[h!]
    \centering
    \setlength{\tabcolsep}{8pt}
    \renewcommand\arraystretch{0.95}
    \adjustbox{width=0.8\textwidth}{\begin{tabular}{c|c|ccc|ccc|ccc}
        \toprule
        \multirow{2}{*}{Model} & \multirow{2}{*}{Augmentation} & \multicolumn{3}{c|}{W$\rightarrow$N} & \multicolumn{3}{c|}{K64 $\rightarrow$ K32} & \multicolumn{3}{c}{K64 $\rightarrow$ K16} \\
         &  & Car  & Ped  & Cyc & Car  & Ped  & Cyc & Car  & Ped  & Cyc  \\ \hline
        \multirow{7}{*}{PVRCNN} & No Aug & 20.36 & 5.79 & \textbf{0.53} & 77.62 & 53.96 & 50.3 & 54.07 & 27.13 & 26.36 \\ 
         ~ & GT-Sampling & 15.86 & 5.29 & 0.0 & 77.97 & 47.83 & 60.54 & 57.61 & 14.45 & 25.44 \\
         ~ & Mixed GT-Sampling & 20.57 & 7.62 & 0.0 & 78.79 & 55.37 & 50.74 & 55.92 & 23.93 & 27.23\\
         ~ & Shape Augmentation (SA) & 20.16  & 5.73 & 0.0 & 78.46 & 54.98 & 56.11 & 59.08 & 33.74 & 30.45 \\ 
         
        ~ & Line Downsampling (LD) &\textbf{ 23.97} & \textbf{10.57} & 0.0 & \textbf{82.72} & \textbf{61.28} & \textbf{64.24} & \textbf{71.57} & \textbf{51.64} & \textbf{46.33}\\

        
        & \textit{Oracle} & 37.85 & 24.56 & 1.67 & 81.45 & 51.75 & 61.55 & 72.71 & 53.05 & 49.8 \\ 
        

        \bottomrule
        
    \end{tabular}}
    \caption{Impact of common and introduced data augmentations on the OOD performance in high-to-low resolution domain gaps. SA and LD are found to consistently improve the AP on target domains, while the widely used GT-Sampling causes overfitting on some classes.}
    \label{tab:aug_sensor_gap}
    \vspace{-1em}
\end{table*}

\subsection{Data Augmentations}
In \cref{tab:aug_sensor_gap}, we investigate the influence of data augmentations on model robustness across the three benchmarks in the High-to-Low domain gap. In particular, Groundtruth-Sampling (GT-sampling) is a widely used technique that augments the number of foregrounds by drawing samples from a database of source domain objects. We also evaluate shape augmentation (SA) proposed in \cite{sessd}, which removes certain parts of the object and/or downsamples its points. For more investigations, we introduce two other augmentations (see \cref{fig:vis_aug}). (1) Line downsampling (LD), inspired by \cite{beamdistillation}, consists in downsampling Lidar beams of the source scans in the vertical direction by a factor of $2$ with a probability $p_{d=2} = 0.3$ and by a factor of $4$ with a probability $p_{d=4} = 0.2$. Unlike \cite{beamdistillation}, no teacher-student training is needed. (2) A variant of GT-sampling, which we term Mixed GT-Sampling, consists in adding samples extracted from another dataset with a different V-FOV compared to the source dataset. In this case, we add samples from KITTI for W$\rightarrow$N and samples from NuScenes for the KITTI benchmarks. This technique can also be used with pointclouds synthesized by generative models like~\cite{deepgm, pugan}.

We find that: 1) \textit{GT-Sampling has negative effects on some classes}, as the detection accuracy deteriorates on pedestrians in K64$\rightarrow$K32-16 and cars in W$\rightarrow$N. We hypothesize this is due to overfitting on source domain shapes. 2) \textit{Mixed GT-Sampling can restore the performance drop of GT-Sampling} and adds small gains, reducing the overfitting of the original method. 3) \textit{The impact of point sparsity augmentations on robustness is the most significant}. SA and LD serve as potent domain randomization tools that enhance performance on the sparser target. The mAP can even exceed the PVRCNN oracle on K32. It is worth noting that in the K64$\rightarrow$K32-16 benchmarks, LD is employed to perfectly mimic the target domain, which accounts for the outstanding performance. However, this controlled experiment shows that if point sparsity arises from a different number of beams while maintaining the same V-FOV, this straightforward technique will enhance the model's robustness. It is also the highest performing on the more challenging W$\rightarrow$N.

\section{Weather Domain Gap}
\label{sec:weather}
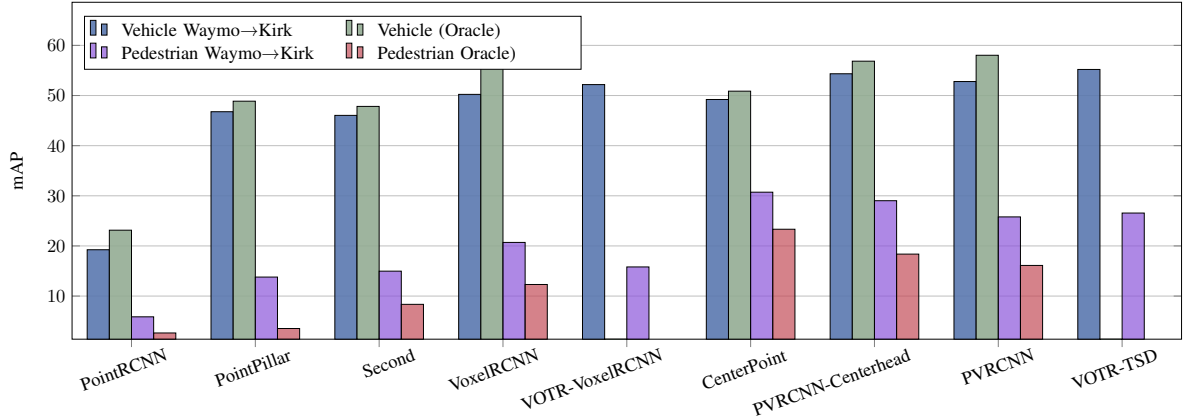
\begin{figure*}[th!]  
\centering
\begin{adjustbox}{width=0.9\textwidth}
\begin{tikzpicture}

\begin{axis}[
    ybar,
    width=1.3\textwidth,
    height=8cm,
    enlargelimits=0.06,
    legend style={
        at={(0.235,0.8)},
        anchor=south,
        legend columns=2,
        column sep=0.2cm,
        text width=3.7cm,
        /tikz/every even column/.append style={column sep=0.5cm} 
    },
    ylabel={mAP},
    ylabel style={yshift=-5pt},
    symbolic x coords={PointRCNN,PointPillar,Second,VoxelRCNN,VOTR-VoxelRCNN,CenterPoint,PVRCNN-Centerhead,PVRCNN,VOTR-TSD},
    xtick=data,
    xtick pos = bottom,
    x tick label style={rotate=20,anchor=east,xshift=20,yshift=-10},
    width=1.3\textwidth,
    bar width=12pt, 
    ymajorgrids=true,
    ymin=5,
    ymax=65,
]
\addplot+[bar shift=-18pt, draw=black, fill={rgb,255:red,90; green,120; blue,175}, fill opacity=0.9] coordinates {(PointRCNN,19.24) (PointPillar,46.77) (Second,46.04) (VoxelRCNN,50.23) (VOTR-VoxelRCNN,52.18) (CenterPoint,49.21) (PVRCNN-Centerhead,54.34) (PVRCNN,52.79) (VOTR-TSD,55.2)};

\addplot+[bar shift=-6pt, draw=black, fill={rgb,255:red,147; green,172; blue,147}, fill opacity=0.9] coordinates {(PointRCNN,23.14) (PointPillar,48.87) (Second,47.83) (CenterPoint,50.87) (VoxelRCNN,57.09) (VOTR-VoxelRCNN, 0) (PVRCNN,58.03) (PVRCNN-Centerhead,56.85) (VOTR-TSD, 0)};

\addplot+[bar shift=6pt, draw=black, fill={rgb,255:red,140; green,85; blue,219}, fill opacity=0.7] coordinates {(PointRCNN,5.86) (PointPillar,13.79) (Second,14.97) (CenterPoint,30.72) (VoxelRCNN,20.7) (VOTR-VoxelRCNN,15.81) (PVRCNN,25.79) (PVRCNN-Centerhead,29.02) (VOTR-TSD,26.56)};

\addplot+[bar shift=18pt, draw=black, fill={rgb,255:red,196; green,76; blue,88}, fill opacity=0.7
] coordinates {(PointRCNN,2.64) (PointPillar,3.53) (Second,8.35) (CenterPoint,23.33) (VoxelRCNN,12.31) (PVRCNN,16.11) (PVRCNN-Centerhead,18.37)};

\legend{Vehicle Waymo→Kirk, Vehicle (Oracle), Pedestrian Waymo→Kirk, Pedestrian Oracle)}
\end{axis}

\end{tikzpicture}
\end{adjustbox}
\vspace{-1em}
\caption{Evaluation of all models on the Waymo$\rightarrow$Kirk benchmark, with Oracle included. The mAP is reported for cars and pedestrians since Kirk has no cyclist class. Surprisingly, the source-only model performs better than the oracle itself, especially on the pedestrian class, revealing the challenge of simple supervised learning on bad weather data.}
    \label{fig:weather_gap_understanding}
    \vspace{-1.5em}
\end{figure*}
The weather domain gap in Lidar is particularly hard to address: it introduces artifacts such as missing points or clutters resulting from light scattering. Sometimes a significant part of the pointcloud can be missing~\cite{spg}. We choose to perform experiments on the Waymo to Kirk benchmark similar to SPG~\cite{spg} instead of introducing simulated weather effects on existing datasets. 

\subsection{Transferability versus Discriminability}
Previous research~\cite{spg} has demonstrated a significant decrease in performance on bad weather data. This has often been formulated as a transferability problem: how can the models trained on good weather data transfer to bad weather data? However, we notice that a simple \textit{in-domain} evaluation of common 3D detectors on bad weather data is missing. In Figure~\ref{fig:weather_gap_understanding}, we assess different models using the Kirk validation split. Each model is trained two times, once on Waymo data (W$\rightarrow$Kr) and once on the Kirk training split, which is regarded as the oracle performance or upper bound. Interestingly, \textit{we find that models trained on clean weather data exhibit comparable or even superior generalization to poor weather conditions compared to those trained on bad weather data}. This is especially true for the class Pedestrian, which exhibits the most deterioration in adverse weather. This finding aligns with a previous work~\cite{rethinkingweather} but takes it a step further: while \cite{rethinkingweather} demonstrates that training on abundant clear weather data (source) is better than training on mixed weather samples (source + target), we show that it is even better than training on a large dataset of target domain only samples. Note that we use a much larger dataset of bad weather conditions than \cite{rethinkingweather}. This reveals that detection in bad weather conditions is not merely a transferability problem; rather, it is a discriminability problem where simple supervised learning is hard. We believe more research should focus on building better models for this domain or on integrating other modalities (camera, radar) to improve detection in adverse conditions.  

\subsection{Data Augmentations and Voxel Size}
In \cref{tab:waymo_augmentation}, we report the result of different design choices using PVRCNN. We find that: 1) \textit{The weather domain gap is not significantly affected by augmentations}. GT-Sampling shows modest gains on the pedestrian class, while the other augmentations slightly increase the vehicle class. 2)\textit{ PVRCNN with a smaller voxel size trained only on $20\%$ of Waymo dataset is able to outperform  SPG~\cite{spg} on the Vehicle class}, as SPG uses a larger voxel size. This highlights again the importance of choosing the right voxel size as it can increase or decrease the model's robustness. 3) \textit{Training on source and target does not improve the performance.}
\begin{table}[h!]
\centering

\resizebox{\linewidth}{!}{
\begin{tabular}{l|cc|cc}
\toprule
\multirow{2}{*}{Method} & \multicolumn{2}{c|}{Waymo} & \multicolumn{2}{c}{Waymo $\rightarrow$ Kirk}\\
& Veh. & Ped.  & Veh. & Ped.  \\ \midrule
No Augmentation w/ Voxel (0.1, 0.1, 0.15)  & 71.22 & 67.72  &  55.48 & 22.20\\
\midrule
GT-Sampling      & 71.01  &  \textbf{68.23} & 55.02  & 23.85 \\
GT-Sampling + LD  & 71.22& 67.62 & \textbf{56.07} & 22.04  \\
GT-Sampling + SA & \textbf{71.66} & 68.11 & 55.72 & 21.48 \\ 
Mixed Domain training (source + target) & 70.58 & 65.91 & 54.24 & 21.34\\
\midrule
SPG$^{*}$ w/ Voxel (0.2, 0.2, 0.3) ~\cite{spg} & 70.63 & 62.31 & 53.51 & \textbf{26.44} \\
No Augmentation w/ Voxel (0.1, 0.1, 0.1)  & 71.27 & 68.21 & 55.78 & 23.73\\
\bottomrule 
\end{tabular}}
\caption{Results of different augmentation and voxel size setttings on Waymo $\rightarrow$ Kirk using PVRCNN. $^{*}$ trains on the whole Waymo dataset. No setting is optimal, but a small voxel size can outperform SPG on the class Vehicle, even though no DA method is used. }
\label{tab:waymo_augmentation}
\vspace{-1.5em}
\end{table}
\section{Discussion and Conclusion}
\label{sec:conclusion}
We explored the robustness of Lidar-based 3D detectors across three main domain gaps: sensor, weather, and location. We evaluated nine state-of-the-art models on six common Domain Adaptation benchmarks. By conducting large-scale experiments and observing various design choices in the 3D-OD pipeline - architecture, data augmentation, anchor size, and voxel size - we draw several important findings across this pipeline. 

\noindent\textbf{Anchor Size}
\begin{enumerate}
    \item Changing anchor size at test-time can notably enhance model performance across various locations, indicating that contrary to common belief, models have learned a generalizable representation, albeit at different scales. 
    \item Training with larger anchors is recommended for practical applications, as the optimal anchor is always found to be smaller than the training anchor. This can enable semi-supervised, federated, and test-time DA without model retraining, allowing a seamless transition between locations by only changing the anchor. 
\end{enumerate}

\noindent\textbf{Architecture and Voxel Size} 
\begin{enumerate}
    \item Point-voxel representations in the backbone show better robustness across most domain pairings. 
    \item Voxel transformer backbones are more robust than 3D CNNs when coupled with point features. 
    \item Anchorless detectors are the most robust in adverse weather.
    \item While shorter voxel heights generally improve performance on the source domain, increasing voxel height improves the robustness when training on high-resolution sensors and inferring on low-resolution sensors.
\end{enumerate}

\noindent\textbf{Data and Augmentations}
\begin{enumerate}
    \item While ground-truth sampling is known to enhance performance on the source domain, we find it has negative effects on the robustness on lower-resolution sensors, likely due to overfitting on the source domain shapes. Adding groundtruth samples from another dataset with a different V-FOV reduces the overfitting.
    \item Point sparsity augmentations, like shape and line downsampling augmentations, prove effective on high-to-low resolution domain gaps. Differences in the V-FOV remain the most challenging. 
    \item Going from high-to-low resolution is more challenging than the reverse due to point sparsity at test-time, with models performing better on higher-resolution test data.
    \item Surprisingly, training on clean weather samples leads to more robustness on bad weather than direct training on bad weather or mixed samples. This highlights the importance of choosing the right training data but also points out that the problem with bad weather data is about discriminability rather than transferability.  
\end{enumerate}

\noindent While our study is exhaustive, there is further work to be done. We have shown that simple architecture and data tricks can improve robustness. Yet, a universally adaptive model remains elusive. We hope our benchmarks and in-depth analysis can benefit the 3D-OD and DA communities. 



\clearpage
\clearpage
{
   \small
   \bibliographystyle{ieeenat_fullname}
   \bibliography{main}
}

\clearpage
\setcounter{page}{1}
\maketitlesupplementary

In the supplementary material, we provide more details and experiments.
\begin{itemize} 
    \item \cref{sec:implementation}: implementation details for the conducted experiments.
    \item \cref{sec:rw3DoD}: related works for lidar-based 3D-OD.
    \item \cref{sec:location_more}: further experiments concerning the location domain gap. This includes the effects of the voxel size, the impact of various anchor sizes, the optimization procedure, and single-model-multiclass experiments.
    \item \cref{sec:sensor_more}: further experiments on the effect of augmentations on VOTR-TSD and SECOND. 
    \item \cref{sec:tpfpfn}: Analysis of True Positives, False Positives and False Negatives per model and dataset
\end{itemize}

\section{Implementation details}
\label{sec:implementation}
For all experiments, we have used the OpenPCDet repository~\cite{openpcdet} and the official implementation of VOTR-TSD~\cite{votrgithub}. For all models, we use a batchsize of 8 to fit on our GPUs. To study the isolated effects of the architectural design choices, voxel encoding, and anchor size, we train the models without GT-Sampling.

\section{Related Works: 3D Object Detection}
\label{sec:rw3DoD}
Lidar-based 3D Object Detection can be divided into three groups: point-based, range image-based, and voxel-based. Point-based methods~\cite{frustum, pointrcnn, 3dssd} extract 3D structural features from the raw points directly using permutation invariant feature extractors like Pointnet~\cite{pointnet, pointnet++}. Range image detectors project the pointcloud onto a 2D plane using spherical projection and employ 2D CNNs for detection~\cite{rangedet, rsn, rangedilated}. Voxel-based methods divide the pointcloud into regular voxels and encode the points inside the voxels using point operations; then, they employ 3D and 2D backbones to generate 3D boxes. VoxelNet~\cite{voxelnet} is a seminal work that implements 3D convolutions on the pointclouds. SECOND~\cite{second} is a first-stage detector that introduces 3D sparse convolutions to boost the efficiency of 3D backbones. CIA-SSD~\cite{ciassd} and SE-SSD~\cite{sessd} both build upon SECOND: CIA-SSD adds IoU prediction to the total loss and uses the predicted IoU values to correct classification scores prior to Non-Maximum Suppression (NMS), while SE-SSD employs a teacher-student framework with diverse shape augmentation strategies to boost the network's capacity of detecting different object shapes. PointPillars~\cite{pointpillars} encodes the pointcloud into bird's eye view (BEV) features and uses a 2D network on this representation. Centerformer~\cite{centerformer} is a one-stage anchorless detector that leverages a DETR~\cite{detr} transformer in its detection head.  There have also been many two-stage approaches that exploit voxel operations using 3D convolutions in the backbone like \cite{pvrcnn, voxelrcnn, points2parts, pyramidrcnn} or 3D transformers like \cite{votr, sst}.  PV-RCNN~\cite{pvrcnn}, Pyramid-RCNN~\cite{pyramidrcnn} and VOTR-TSD~\cite{votr} use point features in the detector to refine the proposals, while VoxelRCNN uses pure voxel-level features eliminating the substantial computational overhead of point-level features. CenterPoint~\cite{centerpoint} presents an anchorless two-stage detector. BtcDet~\cite{btcdet} predicts objects' occupancy in the occluded areas and leverages the occupancy to refine the proposals. M3DETR~\cite{m3detr} extracts features from different representations (voxels, points, BEV) and computes the relationships between these features using transformers before using standard detection heads. In this work, we focus on voxel-based methods since they are getting increasingly more attention and have demonstrated high performance on standard detection benchmarks like KITTI~\cite{kitti} and Waymo~\cite{waymo}. Moreover, works on domain adaptation typically employ voxel-based methods, which motivates a closer inspection of these frameworks.

\section{Location Domain Gap: Further Experiments}
\label{sec:location_more}

\begin{table}[!t]
    \centering
    \adjustbox{width=\linewidth}{\begin{tabular}{l|l|c|c|c}
    \hline
        Model & Voxel Size & Waymo  & W $\rightarrow$ K & W $\rightarrow$ N \\ \hline
        \multirow{4}{*}{PVRCNN} & [0.1, 0.1, 0.125] & \textbf{62.95} &  16.25 &  18.81\\ 
        ~ & [0.1, 0.1, 0.15] &  62.60 & \textbf{16.41} & 18.69 \\ 
        ~ & [0.1, 0.1, 0.2] &  62.41 &  16.33 & 19.51 \\ 
        ~ & [0.1, 0.1, 0.25] & 62.32 &  15.94 & \textbf{19.83} \\ 
        \midrule
        \multirow{4}{*}{VOTR-TSD} & [0.1, 0.1, 0.125] & 64.69 &  \textbf{16.97} &  19.85 \\ 
        ~ & [0.1, 0.1, 0.15] & \textbf{65.20} & 14.46 & \textbf{21.66} \\ 
        ~ & [0.1, 0.1, 0.2] & 64.09 & 15.35 &  20.59 \\ 
        ~ & [0.1, 0.1, 0.25] &  64.43 &  16.86 & 20.43
        \\ \hline
    \end{tabular}}
    \caption{Impact of the voxel size on the out-of-domain performance across different locations. The Car 3D-AP is reported for PVRCNN and VOTR-TSD.}
    \label{tab:env_voxel_encoding}
    \vspace{-1em}
\end{table}
\noindent\textbf{Voxel Size.}  In ~\cref{tab:env_voxel_encoding}, we explore the effect of the voxel size on performance across multiple geographical locations. No significant effect is observed on W$\rightarrow$K. This can be attributed to the primary domain shift type being the object size rather than point sparsity, which renders the anchor size the primary performance driver at test time. However, on W$\rightarrow$N, some performance gain is observed when choosing a higher voxel height, as NuScenes has a lower resolution than Waymo.  
\begin{table}[!ht]
    \centering
    \adjustbox{width=\linewidth}{\begin{tabular}{llll}
    \hline
        Models & Car & Ped & Cyc \\ \hline
        PointRCNN & 5.04 / 38.15 & 28.22 / 31.76 & 0 / 0 \\ 
        PointPillars & 12.75 / 66.27  & 48.34 / 49.47 & 34.9 / 38.51 \\ 
        Second & 9.91 / 65.71 & 41.39 / 46.09 & 22.74 / 25.37 \\ 
        VoxelRCNN & 20.09 / 59.70 & 55.33 / 59.82 & 34.81 / 40.79 \\ 
        VOTR-VoxelRCNNhead & 21.34 / 55.91 & 29.96 / 28.09 & 1.63 / 3.55 \\ 
        PVRCNN & 16.61 / 55.99 & 50.22 / 56.31 & 34.09 / 34.02 \\ 
        VOTR-TSD & 15.75 / 59.39 & 46.19 / 48.11 & 39.28 / 46.82 \\ \hline
    \end{tabular}}
    \caption{Benchmarking anchor-based architectures on W$\rightarrow$K. We report the performance before and after anchor optimization for all classes and denote a consistent improvement for all models.}
    \label{tab:anchor_multiclass}
\end{table}
\begin{table}[!ht]
    \centering
    \begin{tabular}{ccc|c}
    \hline
        \multicolumn{3}{c|}{Anchor Size} & BEV / 3D AP \\ \hline
        3.9 & 1.6 & 1.56 & 52.80 / 11.05 \\ 
        \hline
        3.8 & 1.6 & 1.56 & 61.61 / 16.24 \\ 
        3.7 & 1.6 & 1.56 & 66.17 / 21.58 \\ 
        \textbf{3.6} & 1.6 & 1.56 & 68.76 / 24.63 \\ 
        3.5 & 1.6 & 1.56 & 68.90 / 23.70 \\ 
        3.4 & 1.6 & 1.56 & 68.02 / 20.55 \\ 
        \hline
        3.6 & 1.5 & 1.56 & 79.26 / 49.86 \\ 
        3.6 & 1.4 & 1.56 & 82.11 / 63.49 \\ 
        3.6 & \textbf{1.3} & 1.56 & \textbf{82.50} / 65.03 \\ 
        3.6 & 1.2 & 1.56 & 80.38 / 60.18 \\
        \hline
        \textbf{3.6} & \textbf{1.3} & \textbf{1.5} & 81.38 / \textbf{68.19} \\ 
        3.6 & 1.3 & 1.4 & 80.45 / 58.01 \\ 
        3.6 & 1.3 & 1.6 & 82.53 / 60.88 \\ \hline
    \end{tabular}
    \caption{Illustration of the anchor optimization procedure on the SECOND model on Waymo$\rightarrow$KITTI. The first row denotes the default training and testing anchor size.}
    \label{tab:anchor_optimization}
    \vspace{-1em}
\end{table}

\noindent\textbf{Multiclass Experiments.} In \cref{tab:anchor_multiclass}, we benchmark all anchor-based architectures on the W$\rightarrow$K benchmark, reporting the 3D-AP for all classes. The results show an increase in performance for all models across all classes when tuning the test-time anchor size. This is more pronounced in cars and cyclists than pedestrians, as these classes exhibit a more significant change in size across datasets than pedestrians. Note the results of the class car are slightly different from \cref{table:anchor_all_models}, where all models were trained only on the class car. 

\noindent\textbf{Anchor Size Optimization Procedure.} The anchor size optimization procedure is shown in~\cref{tab:anchor_optimization} on the SECOND model for Waymo $\rightarrow$ KITTI. We adopt a greedy-like optimization approach, which involves altering a single size dimension while maintaining the others constant. Subsequently, the dimension that improves the BEV/3D-AP the most is fixed, and the process is repeated for the next dimension until all three dimensions have been optimized. Our findings reveal a notable relationship between the length and width of the anchors and the overall detection efficacy in this specific benchmark. Notably, adjustments to the width dimension have a pronounced effect on the OOD performance. This observation suggests that objects in the KITTI dataset are generally narrower compared to those in Waymo, implying that smaller anchor widths are more suitable for the KITTI objects. However, the determination of which dimension is most crucial varies depending on the size characteristics of objects in the source and target datasets. For test-time applications, we select the optimal anchor size, which is emphasized in bold in the table.

\noindent\textbf{Tuning the Anchor Size on Target Datasets with Larger Objects.} In \cref{table_anchor_NTK}, we report the effect of changing the anchor size of SECOND on N$\rightarrow$W, where the target objects in this setting are larger than the source objects. We find that the best-performing anchors are still smaller than the training anchors. A larger training anchor can provide better source-only performance than smaller training anchors and allows for an even higher performance at test-time after tuning the anchor size. This shows the importance of choosing a large training anchor size, which sets a wide enough spatial prior for object detection at different sizes. Nevertheless, the enhancements in this domain gap are not as significant as those observed in the transition from Waymo to KITTI. This is primarily because the average object sizes in both datasets are close, and the primary difference in the domain gap is attributed to the resolution and vertical field-of-view.  
\begin{table}[!t]
    \centering
    \adjustbox{width=0.8\linewidth}{\begin{tabular}{lc}
    \hline
        Test-time Anchor & NuScenes→Waymo \\ 
        \hline
        
        Training [3.9, 1.6, 1.56] &   17.31 / 14.94  \\ 
        Best performing [3.8, 1.6 , 1.4] & 18.13 / 15.49  \\ 
        
        \midrule

        Training [4.2, 2.0, 1.6] & 18.03 / 15.39 \\ 
        Best performing [4.2, 2.1, 1.50] & 22.07 / 18.86 \\ 

        \midrule

        Training [4.80, 2.11, 1.79] & 21.01 / 17.96  \\ 
        Best performing [4.60, 2.11 , 1.70] & 22.15 / 18.93  \\ 

        \bottomrule
        
    \end{tabular}}
    \caption{Anchor Study on NuScenes$\rightarrow$Waymo. LEVEL\_1 and LEVEL\_2 AP are reported for the class Vehicle.}
    \label{table_anchor_NTK}
    \vspace{-1em}
\end{table}

\section{Sensor Domain Gap: Further Experiments}
\label{sec:sensor_more}
In \cref{tab:aug_sensor_further}, we investigate the effect of augmentations on VOTR-TSD and SECOND. We add PVRCNN from \cref{tab:aug_sensor_gap} for comparison purposes. We include the oracle and two domain adaptation models ST3D~\cite{st3d} and Beam-Distillation~\cite{beamdistillation}. We notice that foreground augmentations (SA, GT-Sampling, Mixed GT-Sampling) can provide substantial improvements in some cases but are model-dependent and class-dependent. On the other hand, line downsampling proves to provide the most consistent and highest improvement on this domain gap on all benchmarks and models. It is able to come close to and even sometimes outperform the two presented DA models, highlighting the importance of augmentations during source-domain training.    
 \begin{table*}[t!]
    \centering
    \setlength{\tabcolsep}{8pt}
    \renewcommand\arraystretch{0.95}
    \adjustbox{width=0.8\textwidth}{\begin{tabular}{c|c|ccc|ccc|ccc}
        \toprule
        \multirow{2}{*}{Model} & \multirow{2}{*}{Augmentation} & \multicolumn{3}{c|}{W$\rightarrow$N} & \multicolumn{3}{c|}{K64 $\rightarrow$ K32} & \multicolumn{3}{c}{K64 $\rightarrow$ K16} \\
         &  & Car  & Ped  & Cyc & Car  & Ped  & Cyc & Car  & Ped  & Cyc  \\ \hline

        \multirow{6}{*}{SECOND} & No Aug & 17.84 & 4.44 & 0.22 & 73.50 & 41.11 & 39.10 & 50.71 & 16.63 & 17.58 \\ 
         ~ & GT-Sampling & 17.86 & 5.62 & \textbf{3.68} & 73.50 & 38.96 & 51.04 & 55.01 & 9.69 & 22.87 \\ 
         ~ & Mixed GT-Sampling & 17.28 & 2.36 & 0.03 & \textbf{83.20} & 42.80 & 41.75 & 48.72 & 17.43 & 17.98 \\ 
         ~ & Shape Augmentation (SA) & 17.23  & 4.56 & 0.20  & 72.01 & 45.80 & 43.54 & 53.07 & 24.45 & 21.93 \\  
        ~ & Line Downsampling (LD) & \textbf{20.92} & \textbf{7.64} & 0.60 & 75.37 & \textbf{47.35} & \textbf{53.88} & \textbf{65.30} & \textbf{41.16} & \textbf{37.89}\\ 
        \midrule
        \multirow{3}{*}{SECOND} & ST3D~\cite{st3d} & 20.19 & 5.11 & 3.35 & 61.94 & - & - & 52.17 & - & - \\
        ~ & Beam-Distillation~\cite{beamdistillation} & 22.86 & - & - & 74.33 & - & - & 65.13 & - & - \\
        & \textit{Oracle} & 30.30 & 16.79 & 0.0 & 76.61 & 41.43 & 49.74 & 68.34  & 42.96 & 40.70\\ 

        \midrule
         
         \multirow{6}{*}{PVRCNN} & No Aug & 20.36 & 5.79 & \textbf{0.53} & 77.62 & 53.96 & 50.3 & 54.07 & 27.13 & 26.36 \\ 
         ~ & GT-Sampling & 15.86 & 5.29 & 0.0 & 77.97 & 47.83 & 60.54 & 57.61 & 14.45 & 25.44 \\
         ~ & Mixed GT-Sampling & 20.57 & 7.62 & 0.0 & 78.79 & 55.37 & 50.74 & 55.92 & 23.93 & 27.23\\
         ~ & Shape Augmentation (SA) & 20.16  & 5.73 & 0.0 & 78.46 & 54.98 & 56.11 & 59.08 & 33.74 & 30.45 \\ 
        ~ & Line Downsampling (LD) & \textbf{ 23.97} & \textbf{10.57} & 0.0 & \textbf{82.72} & \textbf{61.28} & \textbf{64.24} & \textbf{71.57} & \textbf{51.64} & \textbf{46.33}\\
        \midrule
        \multirow{3}{*}{PVRCNN} & ST3D~\cite{st3d} & 22.99 & - & - & - & - & - & - & - & - \\
        ~ & Beam-Distillation~\cite{beamdistillation} & 25.63 & - & - & - & - & - & - & - & - \\
        & \textit{Oracle} & 37.85 & 24.56 & 1.67 & 81.45 & 51.75 & 61.55 & 72.71 & 53.05 & 49.8 \\

        \midrule

        \multirow{7}{*}{VOTR-TSD} &  No Aug & 21.32 & 7.0 & 3.48 & 76.29 & 49.78 & 49.76 & 55.61 & 21.71 & 24.38 \\ 
        ~ & GT-Sampling & 20.15 & 8.09 & \textbf{6.34} & 78.5 & 41.81 & \textbf{62.13} & 59.37 & 19.04 & 31.25  \\ 
        ~ & Mixed GT-Sampling & 21.54 & 8.22 & 2.41 & 59.49 & 38.64 & 40.68 & 35.86 &  20.27 & 16.82 \\
        ~ & Shape Augmentation (SA) & 20.26 & 7.15 & 3.43 & 75.5 & 51.81 & 50.44 & 57.36 & 29.21 & 27.84 \\
        ~ & Line Downsampling (LD) & \textbf{24.82} & \textbf{9.41} & 5.22 & \textbf{80.3} & \textbf{54.81} & 50.26 & \textbf{69.39} & \textbf{49.32} & \textbf{50.88}  \\
        &  Oracle & 38.13 & 22.51 & 2.43 & 82.47 & 50.69 & 66.08 & 73.75 & 53.45 & 50.69  \\ 
        
        \bottomrule
        \end{tabular}}
    \caption{Impact of common and introduced data augmentations on the OOD performance in high-to-low resolution domain gaps. SA and LD are found to consistently improve the AP on target domains, while GT-Sampling deteriorates pedestrian detection.}
    \label{tab:aug_sensor_further}
    \vspace{-1em}
\end{table*}   

 \begin{figure}[th!]
    \vspace{-6pt}
	\centering
	\includegraphics[width=\linewidth]{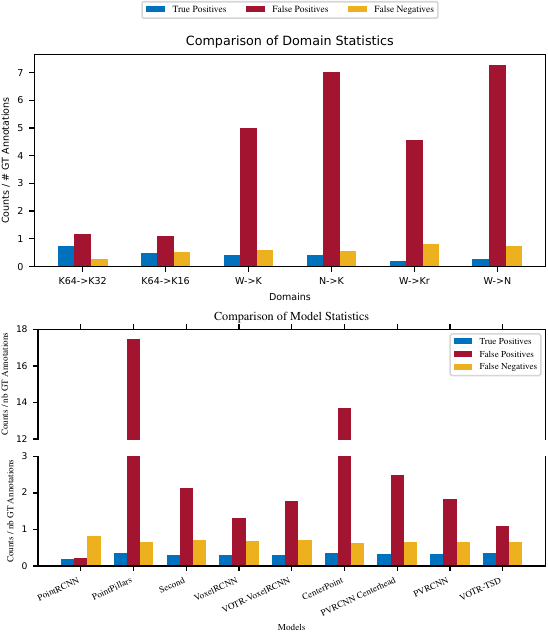}
 \vspace{-12pt}
	\caption{Number of TPs, FPs, and FNs, normalized by the corresponding number of groundtruth annotations.} 
	\label{fig:tp_fp_fn}
 \vspace{-1em}
\end{figure}
\section{TP/FP/FN Analysis}
\label{sec:tpfpfn}
In \cref{fig:tp_fp_fn}, we conduct further analysis, reporting the number of True Positives (TP), False Positives (FP) and False Negatives (FN) for each 3D object model on the six studied benchmarks. The analysis shows: (1) Going from high-to-low resolution results in more FN. The number of FN in K64$\rightarrow$K16 is higher than the number of FN in K64$\rightarrow$K32. The FN ratio is also very high on W$\rightarrow$N and W$\rightarrow$Kr, showing how challenging it is to detect objects when the target domain is sparser than the source domain. (2) Across different geograohical locations, the number of FPs is high. This number can be largely mitigated by tuning the anchor size on the target data. (3) Some models are very prone to FP like PointPillars and CenterPoint, which generate up to 17 FPs for every groundtruth label. (4) Point-based model PointRCNN fails to detect many objects, resulting in the largest number of FN among all studied detectors. (5) The number of FNs among detector shows little variations, showing there is still work to be done to improve the detection of objects in sparser target domains. (6) VOTR-TSD shows the smallest number of FPs among the hybrid and voxel-based methods (PointRCNN has a smaller FP ratio, but it is likely because it generates fewwer bounding boxes than the rest of the models, as reflected in the high FN ratio). (7) Adding point features has a different effect on each backbone: when added to VOTR, point features decrease the number of FPs (compare VOTR-VoxelRCNN to VOTR-TSD). When added to 3D CNNs, the number of FPs slightly increase while FNs slightly decrease.

\end{document}